\documentclass[10pt,twocolumn]{article} 
\usepackage{arxivStyle}
\usepackage{times}
\usepackage{graphicx}
\usepackage{amsmath}
\usepackage{amssymb}
\usepackage{url,hyperref}
\usepackage{pifont}

\begin{document}

\title{Improving Deep Learning-based Automatic Cranial Defect Reconstruction by Heavy Data Augmentation: From Image Registration to Latent Diffusion Models}

\author{Marek Wodzinski$^{1,2}$, Kamil Kwarciak$^{1}$, Mateusz Daniol$^{1}$, Daria Hemmerling$^{1}$ \\
\\
$^{1}$Department of Measurement and Electronics \\
AGH University of Kraków, Kraków, Poland \\
wodzinski@agh.edu.pl  \\
$^{2}$ University of Applied Sciences Western Switzerland (HES-SO Valais) \\
Institute of Informatics, Sierre, Switzerland \\
}

\maketitle
\thispagestyle{empty}

\begin{abstract}
\textbf{
Modeling and manufacturing of personalized cranial implants are important research areas that may decrease the waiting time for patients suffering from cranial damage. The modeling of personalized implants may be partially automated by the use of deep learning-based methods. However, this task suffers from difficulties with generalizability into data from previously unseen distributions that make it difficult to use the research outcomes in real clinical settings. Due to difficulties with acquiring ground-truth annotations, different techniques to improve the heterogeneity of datasets used for training the deep networks have to be considered and introduced. In this work, we present a large-scale study of several augmentation techniques, varying from classical geometric transformations, image registration, variational autoencoders, and generative adversarial networks, to the most recent advances in latent diffusion models. We show that the use of heavy data augmentation significantly increases both the quantitative and qualitative outcomes, resulting in an average Dice Score above 0.94 for the SkullBreak and above 0.96 for the SkullFix datasets. Moreover, we show that the synthetically augmented network successfully reconstructs real clinical defects. The work is a considerable contribution to the field of artificial intelligence in the automatic modeling of personalized cranial implants.
}

\end{abstract}

\section{Introduction}

Every year numerous people suffer from events causing cranial damage and require personalized implants to fill the cranial cavity. The process of modeling and manufacturing personalized cranial implants is costly and time-consuming. Nowadays, the process takes up to several days and is associated with considerable costs~\cite{bonda2015recent,ameen2018design,marreiros2016custom}. However, thanks to the recent advances in artificial intelligence (AI), the process may be partially or fully automated by using dedicated deep learning-based reconstruction methods. Nevertheless, the task is challenging because it is important to properly adjust the implant surface shape and thickness, as well as the edge trim and other factors enabling the formation of the bone-implant connection.

\begin{figure*}[!htb]
    \centering
    \includegraphics[width = 0.9\textwidth]{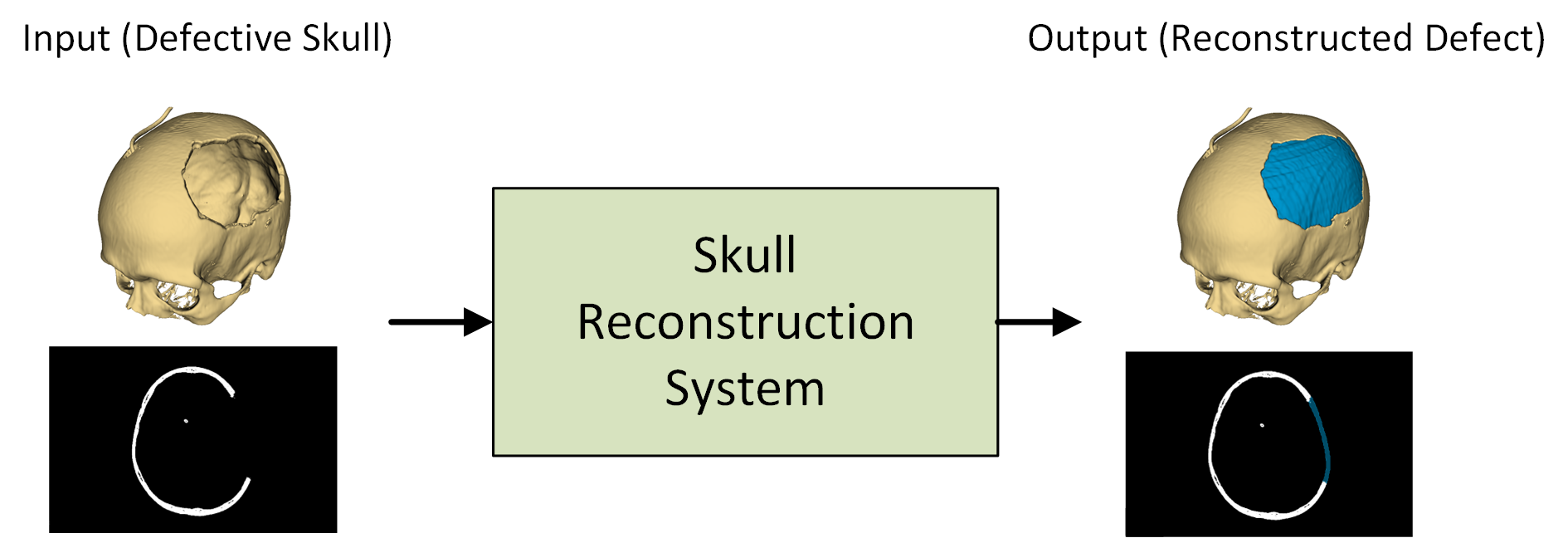}
    \caption{Overview of the defect reconstruction as the volumetric segmentation.}
    \label{fig:overview}
\end{figure*}

The first step of modeling the personalized implant is defect reconstruction. The defect reconstruction may be treated as a volumetric segmentation or shape completion, as shown in Figure~\ref{fig:overview}. However, the problem is considerably challenging for the deep learning methods due to several factors: (i) every skull and defect are different, resulting in enormous heterogeneity, (ii) it is extremely difficult to acquire ground truth for real cranial defects since there are no paired volumes for people before and after the cranial damage, (iii) the skulls are obtained by segmenting high-resolution computed tomography (CT) or magnetic resonance (MR) volumes, resulting in high computational complexity. In this work, we address the first two of these challenges: heterogeneity and the lack of ground truth.

The difficulties with the lack of ground truth for the cranial shape completion motivated researchers to introduce synthetic datasets, SkullBreak and SkullFix, presenting real skulls with artificially introduced cranial defects~\cite{kodym2021skullbreak}. These datasets contain several hundred high-resolution skulls segmented from computed tomography with synthetic defects introduced by morphological operations. Both these datasets were used during the AutoImplant challenge~\cite{li2021autoimplant,li2023towards} that aimed at proposing the most accurate and generalizable solutions. Several notable contributions were proposed~\cite{li2020baseline,kodym2020cranial,kodym2021deep,ellis2020deep,matzkin2020cranial,wodzinski2021improving,jin2020high,pathak2021cranial,mahdi2021u,li2021learning,bayat2020cranial,kroviakov2021sparse,wodzinski2022deep,kwarciak2023deep,li2023sparse}, varying from simple methods based on encoder-decoder networks, to more advanced methods using image registration-based augmentation, sparse convolutional architectures, and shape priors. Nevertheless, all the methods, even the best-performing ones, suffered from difficulties with generalizability into real cranial defects. Even though the methods performed very well on the test cases coming from similar distributions, they had significant difficulties with real clinical cases~\cite{li2023towards}. Recently, researchers attempted to improve the results by reformulating the problem into a point cloud completion task where it is easier to propose highly generalizable solutions, however at the cost of reconstruction accuracy~\cite{sulakhe2022crangan,wodzinski2023high} or processing time~\cite{friedrich2023point}.

In deep learning, the natural way to improve the generalizability of the deep networks is to perform the training data augmentation. The goal of the data augmentation is to synthetically increase the heterogeneity of the training set by randomly modifying existing samples or creating new ones. There are numerous traditional ways to augment the volumetric data, varying from simple flips and crops, through random affine and elastic transformation, to the random intensity distortions and deformable nonrigid registration~\cite{chlap2021review,maharana2022review,perez2017effectiveness}. In the discussed context, since we are interested in augmenting binary data, only the geometric transformations are within the scope of interest. Random geometric augmentation is usually considered an online augmentation technique since it can be applied on the fly due to low computational requirements.

The geometric augmentation techniques, even though useful, are not the only way to increase the generalizability. Another important research area is connected with the use of deep generative networks to generate synthetic samples that can be used to increase the heterogeneity of the training set for a given downstream task~\cite{chlap2021review,antoniou2017data,zaman2020generative,shorten2019survey,chen2022generative}. There are numerous generative architectures, varying from solutions based on variational autoencoders (VAE)~\cite{wang2020data,islam2021crash,papadopoulos2023variational}, deep generative adversarial networks (GAN)~\cite{nie2017medical,frid2018synthetic,bissoto2021gan,huang2018auggan}, to the most recent latent diffusion models~\cite{trabucco2023effective,shivashankar2023semantic,yao2023conditional}. The problem with generative augmentation is mostly connected with the computational complexity of volumetric data and the difficulties with evaluating its medical credibility~\cite{singh2021medical}.

\textbf{Contribution: } In this work we explore several different augmentation possibilities to increase the heterogeneity of training data for the automatic cranial defect reconstruction. We perform large-scale benchmark comparing numerous augmentation methods, varying from the geometric registration, VAEs, and GANs to latent diffusion models and their combinations. Moreover, we perform ablation studies presenting the impact of the synthetic dataset size, regularization coefficient, and sampling strategy. We show that a combination of geometric, registration-based, and generative data augmentation significantly improves the network generalizability, allowing one to train a deep reconstruction network using purely synthetic cases that correctly reconstructs real cranial defects. The combination of heavy geometric augmentation, image registration, and latent diffusion model results in a model outperforming all other state-of-the-art solutions.

\section{Methods}

\subsection{Overview}

We evaluate the following augmentation methods: (i) online geometric augmentation (GA), (ii) offline deformable image registration (IR)~\cite{wodzinski2021improving}, (iii) variational autoencoder (VAE)~\cite{kingma2019introduction,kingma2013auto}, (iv) Wasserstein generative adversarial network with gradient penalty (WGAN-GP)~\cite{gulrajani2017improved}, (iv) vector quantized variational autoencoder (VQVAE)~\cite{van2017neural}, (v) introspective variational autoencoder (IntroVAE)~\cite{huang2018introvae}, (vi) improved introspective variational autoencoder (SoftIntroVAE)~\cite{daniel2021soft}, and (vii) latent diffusion model (LDM) based on VAE and VQVAE~\cite{rombach2022high}.

\begin{figure*}[!htb]
    \centering
    \includegraphics[width = 0.85\textwidth]{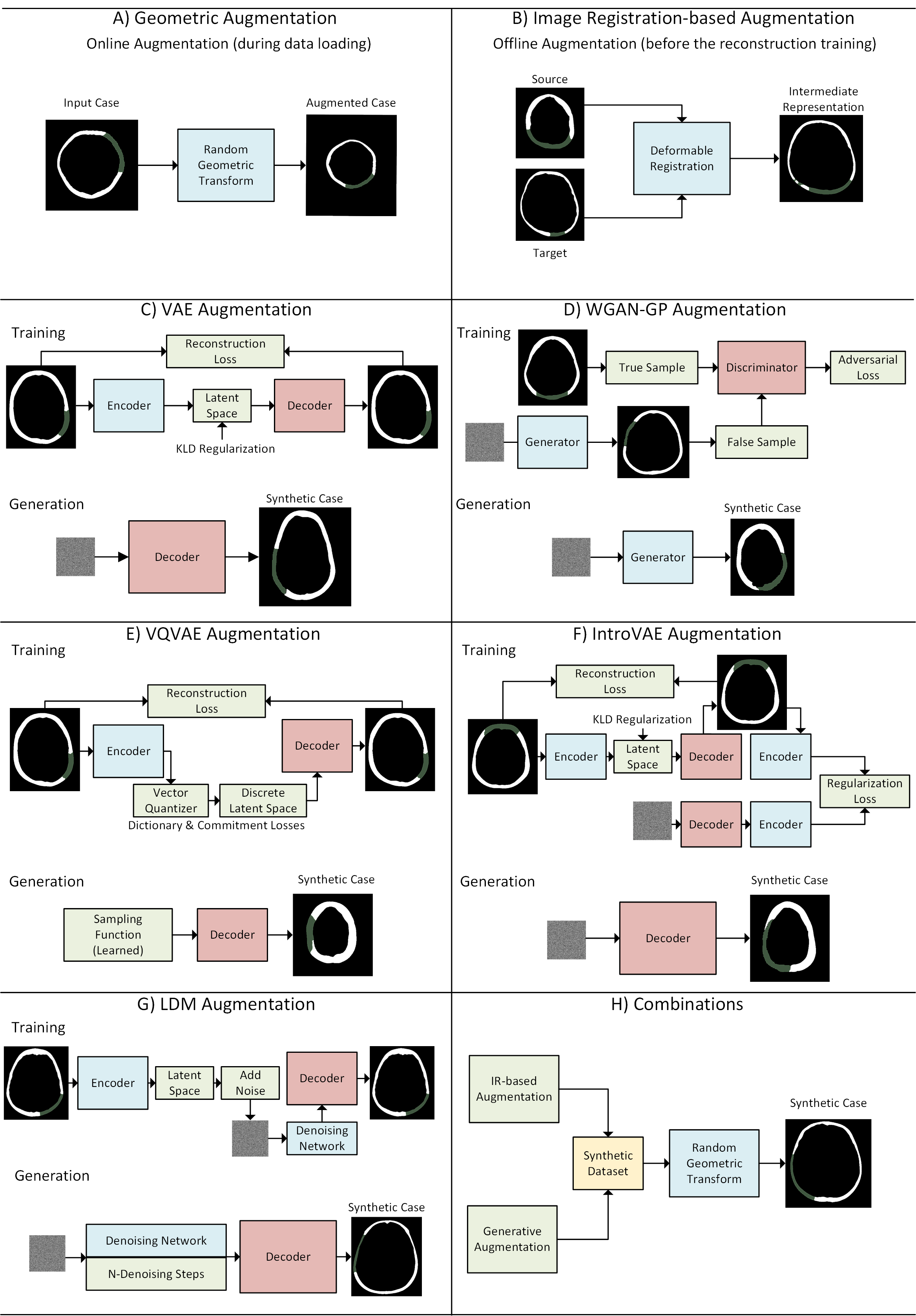}
    \caption{Visualization and comparison of the different augmentation strategies.}
    \label{fig:methods}
\end{figure*}

We decided to use these methods due to several factors. First, previous works presented that the geometric- and registration-based augmentation techniques are considerable solutions to improve the cranial reconstruction quality~\cite{ellis2020deep,wodzinski2021improving}. Second, in this task we are interested in generating binary volumes, therefore contributions focusing on the factors related to the credibility of intensity distributions were not the object of interest. Finally, we train and evaluate the methods using large-scale 3-D volumetric data, resulting in significant computational complexity that forced us to skip several adversarial methods exceeding the available computational resources. The overview of the evaluated methods is shown in Figure~\ref{fig:methods}.

\subsection{Geometric Augmentation}

The evaluated geometric augmentation consists of four types of transformations: (i) random flips, (ii) random crops, (iii) random affine transformation including translation, rotation, and scaling, and (iv) introducing random binary noise. All these transformations are applied with a random probability equal to 0.75, with random order. This kind of augmentation increases the robustness with respect to the relative position, orientation, and scale (which may impact e.g. the robustness of the reconstruction for children). The random binary noise is added to increase the robustness to potential artifacts that may happen during the process of skull segmentation. The visualization of geometric augmentation is shown in Figure~\ref{fig:methods}a.

\subsection{Image Registration}

The registration-based augmentation is based on a multi-level instance optimization deformable registration~\cite{wodzinski2021improving}. The synthetic volumes are generated by randomly choosing two volumes from the training set, considering one as the source, and the second as the target image. The images are sampled to the same resolution and iteratively registered for a given, constant number of iterations, large enough to create an intermediate representation. Since both the registered volumes are binary, the objective function is as follows:
\begin{equation}
    O_{REG}(M, F, u) = MSE(M \circ u, F) + \alpha Reg(u)
\end{equation}
where $M, F$ are the moving and fixed volumes respectively, $MSE$ denotes the mean squared error, $Reg$ is the diffusive regularization, $\alpha$ denotes the regularization coefficient, $u$ is the displacement field, and $\circ$ denotes the warping operation. The visualization of the image registration-based augmentation is shown in Figure~\ref{fig:methods}b.

\subsection{VAE}

The variational autoencoder (VAE) consists of encoder and decoder branches, without the skip connections between the intermediate network levels~\cite{kingma2019introduction}. The output of the last encoder layer is considered to be the embedding. The embedding is represented as a volumetric tensor, without flattening to a feature vector, to maintain the spatial relationships between the learned features. The visualization of the concept is presented in Figure~\ref{fig:methods}c.

The VAE is trained using the following objective function:
\begin{equation}
    O_{VAE}(F, R, E) = Dice(F, R) + \beta KL(E) - \gamma Dice(F_1, F_2), 
\end{equation}
where $F, R$ denote the generated and real volumes respectively, $E$ represents the embedding, $Dice$ is the soft Dice loss, $KL$ denotes the Kullback-Leibler divergence, $F_1, F_2$ denote the channels of the generated volumes (defective skull, defect), and $\beta, \gamma$ control the influence of each term. The KL divergence is used to regularize the latent space while the negative soft Dice loss between the two channels of the generated volume enforces the separation between the generated skull and the defect. The KL divergence is calculated with respect to the standard distribution. During generation, we sampled the standard distribution and propagated it through the decoder.

\subsection{WGAN-GP}

The Wasserstein Generative Adversarial Network with gradient penalty (WGAN-GP) is one of the most successful formulations of GAN-based networks~\cite{gulrajani2017improved}. It consists of a generator responsible for generating synthetic volumes and a discriminator whose goal is to identify between real and synthetic cases. Thanks to the Wasserstein-based formulation and the additional gradient penalty term, the network training is more stable and resistant to mode collapse when compared to the traditional DCGAN. The WGAN-GP method is visualized in Figure~\ref{fig:methods}d.

We trained the WGAN-GP using the following objective function:
\begin{equation}
\begin{split}
    F &= G(z) \\ 
    O_D &= -mean(D(R)) + mean(D(F)) + \lambda GP \\
    O_G &= -mean(D(F)) - \gamma Dice(F_1, F_2),
\end{split}
\end{equation}
where $O_D, O_G$ are the objective functions for the discriminator and the generator, respectively. The $D, G$ are the discriminator and generator networks, $z$ is the latent vector sampled from the standard distribution, $F$ is the generated image, $R$ is the real image, $GP$ denotes the gradient penalty, and $\lambda$ controls the influence of gradient penalty. For reliable comparison, we used the same architecture for the generator as we used for the decoder in VAE. During generation, we sampled the standard distribution and propagated it through the generator. 

\subsection{VQVAE}

The vector quantized VAE (VQVAE) is similar to the vanilla VAE, however, with the difference that the encoder outputs a discrete representation instead of the continuous one~\cite{van2017neural}. Such a representation, when combined with a learnable prior, can generate high-quality and heterogeneous volumes. Unlike VAE, VQVAE does not require the KL divergence to regularize the latent space, therefore allowing the generation of more heterogeneous cases. The VQVAE is visualized in Figure~\ref{fig:methods}e.

We trained the VQVAE using the following objective function:
\begin{equation}
\begin{split}
    O_{VQVAE}(F, R, E) &= Dice(F, R) + \beta CL(E)  \\ 
    &+ \theta DL(E) - \gamma Dice(F_1, F_2) \\
    CL(E) &= mean((E - Q(E))^2) \\
\end{split}
\end{equation}
where $Q$ is the vector quantizer, $CL$ is the commitment loss, $DL$ is the dictionary loss, and $\beta, \theta$ control influence of both the losses. Both $CL$ and $DL$ are calculated during the quantization of the latent vector. We used the same encoder-decoder architecture for VQVAE as for VAE. The difference is only related to the discrete representation of the latent codes and the objective function. However, during generation, we sampled the latent codes using learned convolutional prior, instead of sampling the standard distribution, as performed for the VAE or WGAN-GP-based generation.

\subsection{IntroVAE \& SoftIntroVAE}

The Introspective VAE (IntroVAE) is a contribution combining the advantages of both adversarial training and VAE~\cite{huang2018introvae}. It allows one to use the VAE formulation based on the encoder-decoder architecture, however, the network is simultaneously enhanced by calculating the embedding between real and synthetic samples to improve the distribution of embeddings in the latent space. The IntroVAE is presented in Figure~\ref{fig:methods}f. The objective function is as follows:
\begin{equation}
\begin{split}
    E_R &= Enc(R) \\
    E_F &= Enc(F) \\
    DL(E_R, E_F) &= \beta KL(E_R) \\
    &+ \theta * (ReLU(I_T - KL(E_R)) \\
    &+ ReLU(I_T - KL(E_F))) \\
    O_{IntroVAE_{Enc}}(F, R, E) &= Dice(F, R) + DL(E) \\ 
    &- \gamma Dice(F_1, F_2) \\
    O_{IntroVAE_{Dec}}(E_R, E_F) &= \beta KL(E_r) + \beta KL(E_f)
\end{split}
\end{equation}
where $E_R, E_F$ denote the embeddings related to the real and generated volumes, $DL$ is the distribution loss, $I_T$ is the IntroVAE threshold,  and $\theta$ controls the influence of the introspective step. The encoder and decoder objective functions are optimized simultaneously by two separate optimizers. The limitation of the IntroVAE setup is related to the necessity to manually tune the $I_T$ and $\theta$ to balance between the introspective and variational losses. The generation of new samples during inference is performed by propagating samples from standard distribution through the decoder network. The SoftIntroVAE overcomes the problem of IntroVAE related to the necessity of manual tuning of associated hyperparameters. The training pipeline is improved by using evidence lower bound (ELBO) formulation. Nevertheless, the overall training procedure and processing pipeline remain similar.

\subsection{Latent Diffusion Model}

The latent diffusion models (LDM) are extending the diffusion models in image or volume spaces by performing the diffusion process directly in the latent space~\cite{rombach2022high}. The LDMs start from calculating the latent representation of the input data (e.g. by using VAE or VQVAE) and then they incrementally add the noise to the latent representation. Such a formulation makes it possible to introduce conditioning based on text or visual inputs. However, in this work, we focus on the ability to increase the heterogeneity of the generated representations by training a sampling prior. The concept of LDMs is presented in Figure~\ref{fig:methods}g. In our study, the denoising network is based on a lightweight UNet architecture attempting to denoise the autoencoder's latent tensors. The objective function during training is as follows:
\begin{equation}
\begin{split}
    O_{LDM}(N_{C}, N_{GT}) &= \frac{1}{N}\sum_{1}^{N}(N(i)_{C} - N(i)_{GT})^2,
\end{split}
\end{equation}
where $N_{C}, N_{GT}$ denote the estimated and the ground-truth noise at the given denoising step.

In this work, we decided to evaluate only the LDM formulation, without comparison to the diffusion models trained directly in the volumetric space. The reason for this was connected with the fact that training the volumetric diffusion models required unacceptable computational resources exceeding several weeks of training for a single ablation study. We evaluated the influence of LDMs for both VAE and VQVAE-based autoencoders. The generation of new samples is performed by sampling standard distribution, performing a complete, iterative denoising process, and then propagating the embedding through the decoder to obtain the generated volume.

\subsection{Combinations}

The presented generation or augmentation methods have different properties. Therefore, we decided to evaluate their combinations as well. We combined the online geometric augmentation with offline image registration-based or generative-based augmentation. We also combined the registration-based augmentation directly with the synthetically generated cases because we noted that both methods result in visually different outcomes, leading to a conclusion that their combination may further increase the training set heterogeneity. Geometric augmentation is always applied as the last one after the particular case is loaded or generated.
The concept of combining the augmentation techniques is presented in Figure~\ref{fig:methods}h.

\subsection{Downstream Task - Automatic Defect Reconstruction}

\begin{figure*}[!htb]
    \centering
    \includegraphics[width = 1.0\textwidth]{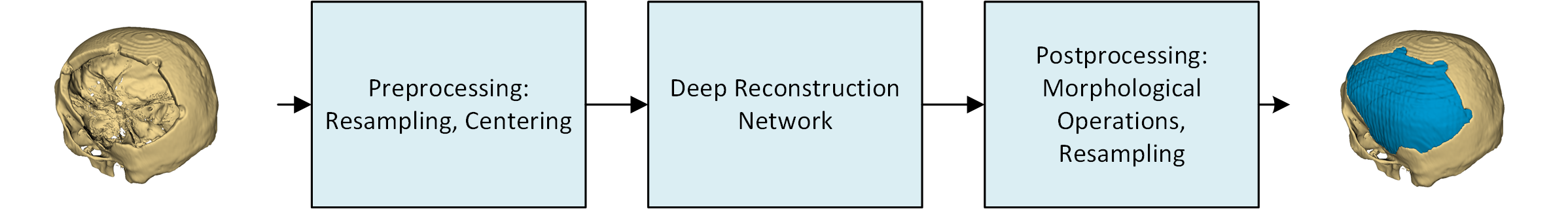}
    \caption{The pipeline of the defect reconstruction process.}
    \label{fig:pipeline}
\end{figure*}

The augmentation methods are evaluated by training a deep convolutional network dedicated to supervised cranial defect reconstruction. The problem may be considered as a volumetric segmentation task. We decided to use the state-of-the-art volumetric residual UNet as the reconstruction network~\cite{wodzinski2022deep}, firstly because of the relatively low computational complexity, and secondly, because our other work presented that there are no significant differences between the UNet or other contributions based on e.g. vision transformers~\cite{wodzinski2023automatic}. The reconstruction network was trained using the Soft Dice Loss as the objective function. The input to the network was the volumetric skull with a defect, and the goal of the network was to directly predict the missing part of the skull. To ensure an unbiased evaluation, all the augmentation methods were evaluated using the same reconstruction architecture. The reconstruction pipeline is shown in Figure~\ref{fig:pipeline}.

All the volumes are preprocessed by centering the defective skull with a predefined offset (distance of the skull to the grid boundary, to accommodate for cases where the defect may be larger than the defective outline). The volumes are resampled to 256x256x256 for the training and inference, however, the evaluation is performed on the original resolution for a particular case. After the inference, the reconstructed output is post-processed by morphological operations that delete all binary structures whose volume is below a certain threshold and that overlap with the defective skull. The same pre- and post-processing methods are applied to all experiments.

\subsection{Datasets}

We use the SkullFix and SkullBreak datasets to train and evaluate the augmentation methods~\cite{kodym2021skullbreak}. The training part of the SkullFix dataset consists of 100 defective skulls and the test dataset has 110 defective cases. The SkullBreak consists of 114 skulls, each with 5 different defects, placed randomly in the bilateral, frontoorbital, and parietotemporal bones, resulting in 570 cases. The test part of the SkullBreak dataset consists of 100 defective cases. Both the SkullFix and SkullBreak datasets consist of real skulls with synthetically introduced defects to create the ground truth. Unfortunately, it is inherently difficult to acquire real ground-truth cases for cranial defect reconstruction because it would require paired CT scans before and after a given traumatic event. 

\begin{figure*}[!htb]
    \centering
    \includegraphics[width = 1.00\textwidth]{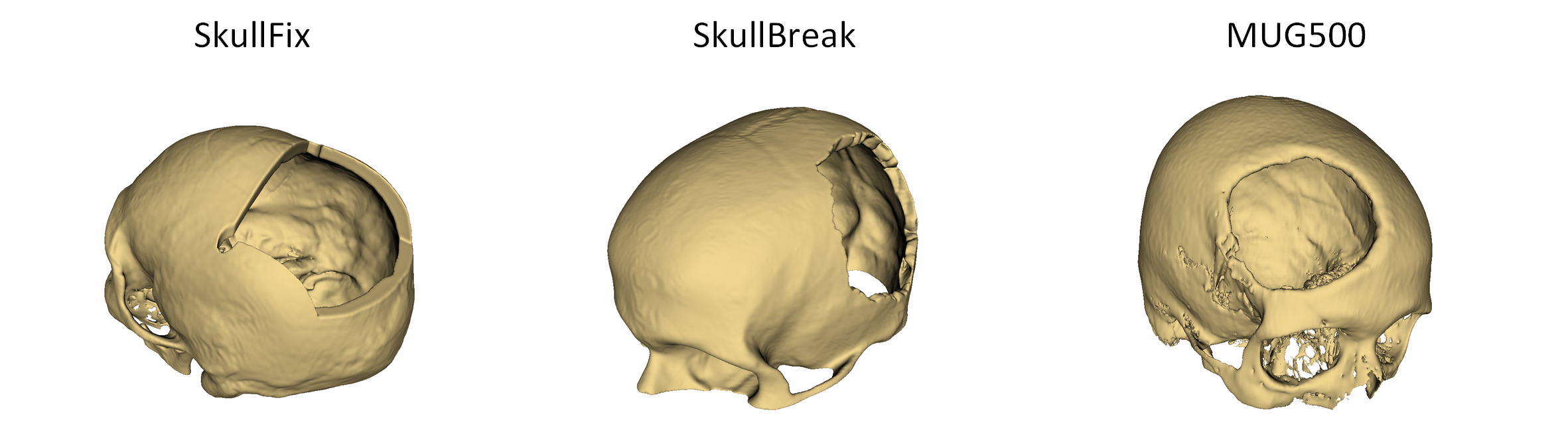}
    \caption{Exemplary skulls from the three datasets: (i) SkullFix~\cite{kodym2021skullbreak}, (ii) SkullBreak~\cite{kodym2021skullbreak}, (iii) MUG500~\cite{li2021mug500}. The SkullFix dataset contains synthetic defects located mostly in similar locations, with partially available facial structures. The SkullBreak represents numerous heterogeneous synthetic defects with various sizes and different locations. The MUG500 dataset contains real cranial defects.}
    \label{fig:datasets}
\end{figure*}

Additionally, we evaluate the proposed method on 11 real clinical cases that were made available for the generalizability sub-task in the AutoImplant challenge and 29 clinical cases from the MUG500 dataset~\cite{li2021mug500}. In contrast to the SkullFix and SkullBreak datasets, the cases present real skulls with real defects. Unfortunately, since the ground truth cannot be defined for such cases, the evaluation can be performed only qualitatively. The missing part of the skull is unknown and there is more than one potential solution for the personalized implant, based on the desired material and other properties. Exemplary visualization of cases from the SkullFix, SkullBreak, and MUG500 datasets are shown in Figure~\ref{fig:datasets}.

\subsection{Experimental Setup}

The experiments for the offline augmentation methods are performed in two steps. First, a particular method is trained and used to generate new synthetic samples. Then, the generated dataset is combined with the training part of the SkullFix and SkullBreak datasets and used for training the reconstruction network for the downstream task. A small subset of the training part of both datasets is used as the validation set to keep track of the training progress. In the case of experiments involving only online augmentation techniques (no augmentation and geometric augmentation), the reconstruction network is trained directly, without the first phase.

We perform ablation studies to verify: (i) the quantitative differences between different augmentation methods, (ii) the impact of synthetic dataset size, (iii) the impact of regularization coefficient for the registration-based augmentation, (iv) the impact of the generative sampling technique for the generative networks, (v) the impact of different strengths of the geometric augmentation.

All the experiments are evaluated quantitatively using the test subset of the SkullFix and SkullBreak datasets, and qualitatively using a dataset consisting of real cranial defects. The test set was not used at all during training the augmentation techniques and the reconstruction network.

All the generative networks were trained until convergence with various hyperparameters setups to find an optimal solution. The training of the generative networks was performed using the PLGrid Athena supercomputer consisting of clusters with NVIDIA A100 GPUs with 40GB dedicated HBM memory. The inference was performed using a workstation equipped with the NVIDIA A6000 GPUs. During inference, the defect reconstruction method does not require more than 12GB of dedicated GPU memory and the reconstruction, including loading, preprocessing, and postprocessing requires (on average) less than 5 seconds. All the processing operations are accelerated using GPU.

To quantitatively evaluate the augmentation techniques we use the Dice Score (DSC), 95th percentile of Hausdorff distance (HD95), surface Dice Score (SDSC), and mean surface distance (MSD). Unfortunately, these metrics cannot be calculated for the real cases (no ground truth available), however, the visual comparison is presented that shows the superior generalizability of the strongly augmented reconstruction methods. Any claim about a statistical improvement is supported by a Wilcoxon signed-rank test with a p-value below 0.01.

\section{Results}

\subsection{Impact of Dataset Size}

\begin{figure}[!htb]
    \centering
    \includegraphics[width = 0.5\textwidth]{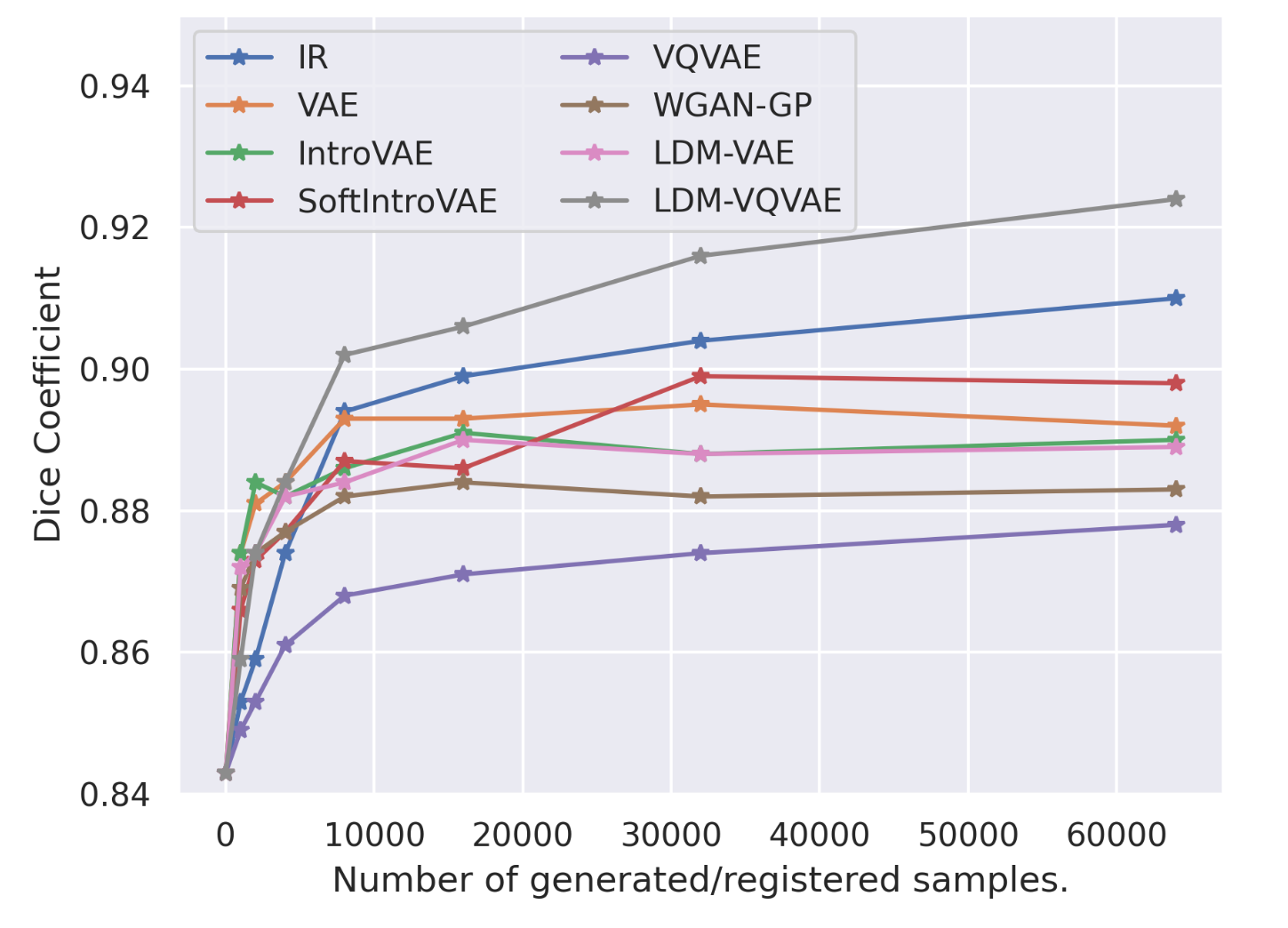}
    \caption{The impact of the number of generated or registered samples on the Dice coefficient evaluated using the SkullBreak test set. It can be noted that the VAE-based methods quickly saturate while the VQVAE and IR continue to improve the generalizability.}
    \label{fig:datasetsize}
\end{figure}

The first experiment shows the impact of the synthetic dataset size on the quality of the following cranial defect reconstruction. The results presenting the impact of the number of synthetic samples for the registration-based and generative augmentation methods are presented in Figure~\ref{fig:datasetsize}.

It can be noted that even a small number of synthetic cases (1000) strongly improves the results compared to the baseline without any augmentation for all the evaluated methods. Further increase in the number of training samples constantly improves the results for the IR-based augmentation, the VQVAE network, and the LDM based on the VQVAE. In contrast, the results for VAE, IntroVAE, SoftIntroVAE, and WGAN-GP quickly saturate and there is no statistically significant difference between 32000 and 64000 synthetic samples. This suggests that methods generating new cases based on sampling the normal distribution tend to generate relatively homogeneous volumes with a lot of similar features.

\subsection{Impact of Geometric Heterogeneity}

Separately from the generative methods, it is interesting to observe the impact of the geometric augmentation strength. The setups of these experiments are presented in Table~\ref{tab:geosetting}. The results presenting the impact of the geometric augmentation with respect to the augmentation strength are shown in Table~\ref{tab:geometric}.

\begin{table}[!htb]
\centering
\caption{Comparison of the geometric augmentation settings. Flips: S - sagittal, F - frontal, L - logitudinal. Affine: D - degrees, V - voxels, S - scale. Noise: S - standard deviation, T: thresholds.}
\renewcommand{\arraystretch}{1.0}
\footnotesize
\resizebox{0.45\textwidth}{!}{%
\begin{tabular}{lcccc}
\label{tab:geosetting}
Setting & Flips & Crops & Affine & Binary Noise \tabularnewline
\hline

Basic & \checkmark (S) & \ding{55} & 7D, 7V, S: 0.7-1.1 & \ding{55} \tabularnewline
\hline

Heavy & \checkmark (S) & \checkmark & 15D, 10V, S: 0.5-1.2 & T: 2.2-4.5, S: 1.0 \tabularnewline
\hline

Extreme & \checkmark (SFL) & \checkmark & 45D, 15V, S: 0.4-1.3 & T: 1.8-4.5, S: 1.0  \tabularnewline
\hline

\hline

\end{tabular}}

\end{table}

Interestingly, the most extreme geometric augmentation improved the results the most, even though the test cases from the SkullBreak and SkullFix datasets are not scaled or rotated. The reason for such a behavior is connected with the overfitting. The reconstruction network without any augmentation or with just minor geometric transformations quickly overfits the training data, limiting the ability to generalize into previously unseen cases. On the other hand, the extreme geometric augmentation that covers flips across all axes, scaling, and rotating makes it difficult for the reconstruction network to overfit the training data. Therefore, for all remaining experiments geometric augmentation refers to the extreme augmentation setting.

\begin{table*}[!htb]
\centering
\caption{Impact of the geometric augmentation settings. The larger the geometric distortions, the better the resistance to overfitting.}
\renewcommand{\arraystretch}{1.0}
\footnotesize
\resizebox{0.99\textwidth}{!}{%
\begin{tabular}{lccccccccc}
\label{tab:geometric}
Geometric Setting & \multicolumn{4}{c}{SkullBreak} & \multicolumn{4}{c}{SkullFix} \tabularnewline
\hline
\multicolumn{1}{c}{} & \multicolumn{1}{c}{DSC $\uparrow$}  &  \multicolumn{1}{c}{SDSC $\uparrow$} & \multicolumn{1}{c}{HD95 [mm] $\downarrow$} & \multicolumn{1}{c}{MSD [mm] $\downarrow$} & \multicolumn{1}{c}{DSC $\uparrow$} & \multicolumn{1}{c}{SDSC $\uparrow$} & \multicolumn{1}{c}{HD95 [mm] $\downarrow$} & \multicolumn{1}{c}{MSD [mm] $\downarrow$} \tabularnewline

Baseline (No Aug) & 0.843 & 0.743 & 2.438 & 0.787 & 0.892 & 0.881 & 1.882 & 0.554 
\tabularnewline

Basic & 0.869 & 0.819 & 1.914 & 0.702 & 0.897 & 0.896 & 1.771 & 0.514 
\tabularnewline

Heavy & 0.872 & 0.817 & 1.879 & 0.672 & 0.901 & 0.913 & 1.682 & 0.501 
\tabularnewline

Extreme & \textbf{0.888} & \textbf{0.824} & \textbf{1.842} & \textbf{0.648} & \textbf{0.904} & \textbf{0.922} & \textbf{1.572} & \textbf{0.482} 
\tabularnewline

\hline

\end{tabular}}

\end{table*}

\subsection{Image Registration - Regularization Coefficient}

An experiment dedicated directly to IR-based augmentation is connected with the impact of the regularization coefficient on the heterogeneity of the resulting samples. A high regularization coefficient prohibits folding and deformations incredible from the medical point of view at the cost of limiting the complexity of the nonrigid deformations. In contrast, a low regularization coefficient may result in implausible deformations of the skull, however modeling complex local deformations. To verify the impact of the regularization coefficient we generated several versions of the IR-based dataset and performed the training. The results presenting the impact of the regularization coefficient are shown in Table~\ref{tab:registration}.

What is interesting, the highly regularized registration result in considerably worse performance on the following defect reconstruction. On the other hand, low regularization coefficients results in numerous folding artifacts, however decreasing the performance only slightly. The best results are obtained with a regularization coefficient equal to 12500 which allows complex deformations and sometimes results in folding. This is an interesting takeaway because it shows that it is more important to increase the heterogeneity than to preserve the quality, at least in the context of data augmentation.

\begin{table*}[!htb]
\centering
\caption{Influence of the regularization coefficient on the performance of the registration-based augmentation. Note that the relatively low regularization coefficient (that allows folding and implausible deformations) improves the reconstruction the most.}
\renewcommand{\arraystretch}{1.0}
\footnotesize
\resizebox{0.99\textwidth}{!}{%
\begin{tabular}{lccccccccc}
\label{tab:registration}
Regularization Coeff & \multicolumn{4}{c}{SkullBreak} & \multicolumn{4}{c}{SkullFix} \tabularnewline
\hline
\multicolumn{1}{c}{} & \multicolumn{1}{c}{DSC $\uparrow$}  &  \multicolumn{1}{c}{SDSC $\uparrow$} & \multicolumn{1}{c}{HD95 [mm] $\downarrow$} & \multicolumn{1}{c}{MSD [mm] $\downarrow$} & \multicolumn{1}{c}{DSC $\uparrow$} & \multicolumn{1}{c}{SDSC $\uparrow$} & \multicolumn{1}{c}{HD95 [mm] $\downarrow$} & \multicolumn{1}{c}{MSD [mm] $\downarrow$} \tabularnewline

Baseline (No Aug) & 0.843 & 0.743 & 2.438 & 0.787 & 0.892 & 0.881 & 1.882 & 0.554 
\tabularnewline

6250 & \textbf{0.911} & 0.903 & 1.612 & 0.529 & \textbf{0.937} & 0.949 & 1.374 & 0.384
\tabularnewline

12500 & 0.910 & \textbf{0.905} & \textbf{1.574} & \textbf{0.511} & 0.934 & \textbf{0.952} & \textbf{1.321} & \textbf{0.362}
\tabularnewline

25000 & 0.907 & 0.901 & 1.599 & 0.522 & 0.931 & 0.949 & 1.352 & 0.371 
\tabularnewline

50000 & 0.905 & 0.882 & 1.642 & 0.548 & 0.926 & 0.941 & 1.371 & 0.389 
\tabularnewline

100000 & 0.901 & 0.867 & 1.781 & 0.572 & 0.919 & 0.932 & 1.394 & 0.394 
\tabularnewline

\hline

\end{tabular}}

\end{table*}

\subsection{Impact of Generative Sampling}

\begin{figure}[!htb]
    \centering
    \includegraphics[width = 0.45\textwidth]{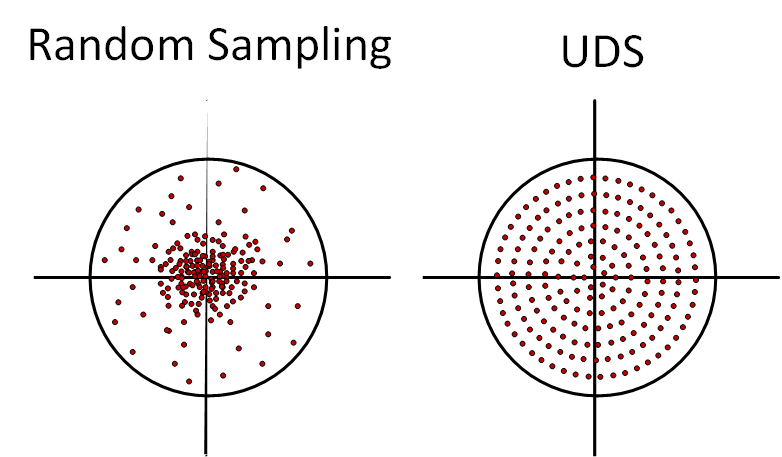}
    \caption{Visual comparison of the two employed latent space sampling strategies. The UDS ensures that the whole latent space is represented and there are not numerous similar generated samples.}
    \label{fig:sampling}
\end{figure}

We performed an experiment dedicated to evaluating the sampling strategy for the networks based on sampling the standard distribution to generate new samples, so the VAE, WGAN-GP, IntroVAE, and SoftIntroVAE. We evaluated three strategies, the first one, the most commonly used, was based on random sampling from the standard distribution. The second strategy was based on random sampling of the uniform distribution. The final one was based on the uniform deterministic sampling (UDS) of the latent space by explicitly and uniformly dividing the N-dimensional sphere with a radius equal to three standard deviations into the desired number of samples. A comparison of these two sampling strategies is shown in Figure~\ref{fig:sampling}.

It can be noted that the UDS improves the results for all the generative networks. The reason behind such results is connected with the fact that the UDS explicitly enforces the heterogeneity of the generated samples, while random sampling tends to generate the majority of the samples close to the center of the distribution. It results in samples that are similar to each other and limit the improvement for the following downstream task.

\begin{table*}[!htb]
\centering
\caption{Comparison of the three sampling strategies (SD - sampling from the standard distribution, UD - sampling from the uniform distribution, UDS - uniform deterministic sampling. Note that UDS improves the outcomes for all methods based on the latent space sampling.}
\renewcommand{\arraystretch}{1.0}
\footnotesize
\resizebox{0.99\textwidth}{!}{%
\begin{tabular}{lccccccccc}
\label{tab:sampling}
Method (Sampling Strategy) & \multicolumn{4}{c}{SkullBreak} & \multicolumn{4}{c}{SkullFix} \tabularnewline
\hline
\multicolumn{1}{c}{} & \multicolumn{1}{c}{DSC $\uparrow$}  &  \multicolumn{1}{c}{SDSC $\uparrow$} & \multicolumn{1}{c}{HD95 [mm] $\downarrow$} & \multicolumn{1}{c}{MSD [mm] $\downarrow$} & \multicolumn{1}{c}{DSC $\uparrow$} & \multicolumn{1}{c}{SDSC $\uparrow$} & \multicolumn{1}{c}{HD95 [mm] $\downarrow$} & \multicolumn{1}{c}{MSD [mm] $\downarrow$} \tabularnewline

Baseline (No Aug) & 0.843 & 0.743 & 2.438 & 0.787 & 0.892 & 0.881 & 1.882 & 0.554 
\tabularnewline

VAE (SD) & 0.874 & 0.851 & 1.942 & 0.582 & 0.902 & 0.908 & 1.561 & 0.439 
\tabularnewline

VAE (UD) & 0.887 & 0.869 & 1.917 & 0.566 & 0.907 & 0.915 & 1.461 & 0.419 
\tabularnewline

VAE (UDS) & \textbf{0.892} & \textbf{0.874} & \textbf{1.906} & \textbf{0.561} & \textbf{0.909} & \textbf{0.917} & \textbf{1.462} & \textbf{0.419} 
\tabularnewline

WGAN-GP (SD) & 0.875 & 0.782 & 2.304 & 0.728 & 0.894 & 0.889 & 1.704 & 0.524 
\tabularnewline

WGAN-GP (UD) & 0.881 & 0.795 & 2.252 & 0.703 & 0.895 & 0.893 & 1.681 & 0.515 
\tabularnewline

WGAN-GP (UDS) & \textbf{0.883} & \textbf{0.797} & \textbf{2.248} & \textbf{0.700} & \textbf{0.897} & \textbf{0.894} & \textbf{1.672} & \textbf{0.512}
\tabularnewline

IntroVAE (SD) & 0.874 & 0.864 & 2.114 & 0.584 & 0.902 & 0.905 & 1.569 & 0.442 
\tabularnewline

IntroVAE (UD) & 0.889 & \textbf{0.882} & \textbf{2.079} & \textbf{0.568} & 0.906 & 0.913 & 1.485 & 0.426 
\tabularnewline

IntroVAE (UDS) & \textbf{0.890} & 0.881 & 2.084 & 0.570 & \textbf{0.908} & \textbf{0.915} & \textbf{1.483} & \textbf{0.421} 
\tabularnewline

SoftIntroVAE (SD) & 0.871 & 0.859 & 2.177 & 0.587 & 0.903 & 0.907 & 1.571 & 0.448 
\tabularnewline

SoftIntroVAE (UD) & \textbf{0.899} & 0.872 & 2.054 & 0.569 & 0.909 & 0.914 & 1.482 & 0.425 
\tabularnewline

SoftIntroVAE (UDS) & 0.898 & \textbf{0.877} & \textbf{2.048} & \textbf{0.566} & \textbf{0.911} & \textbf{0.916} & \textbf{1.479} & \textbf{0.423} 
\tabularnewline

\hline

\end{tabular}}

\end{table*}

\subsection{Augmentation Method}

\begin{figure*}[!htb]
    \centering
    \includegraphics[width = 1.00\textwidth]{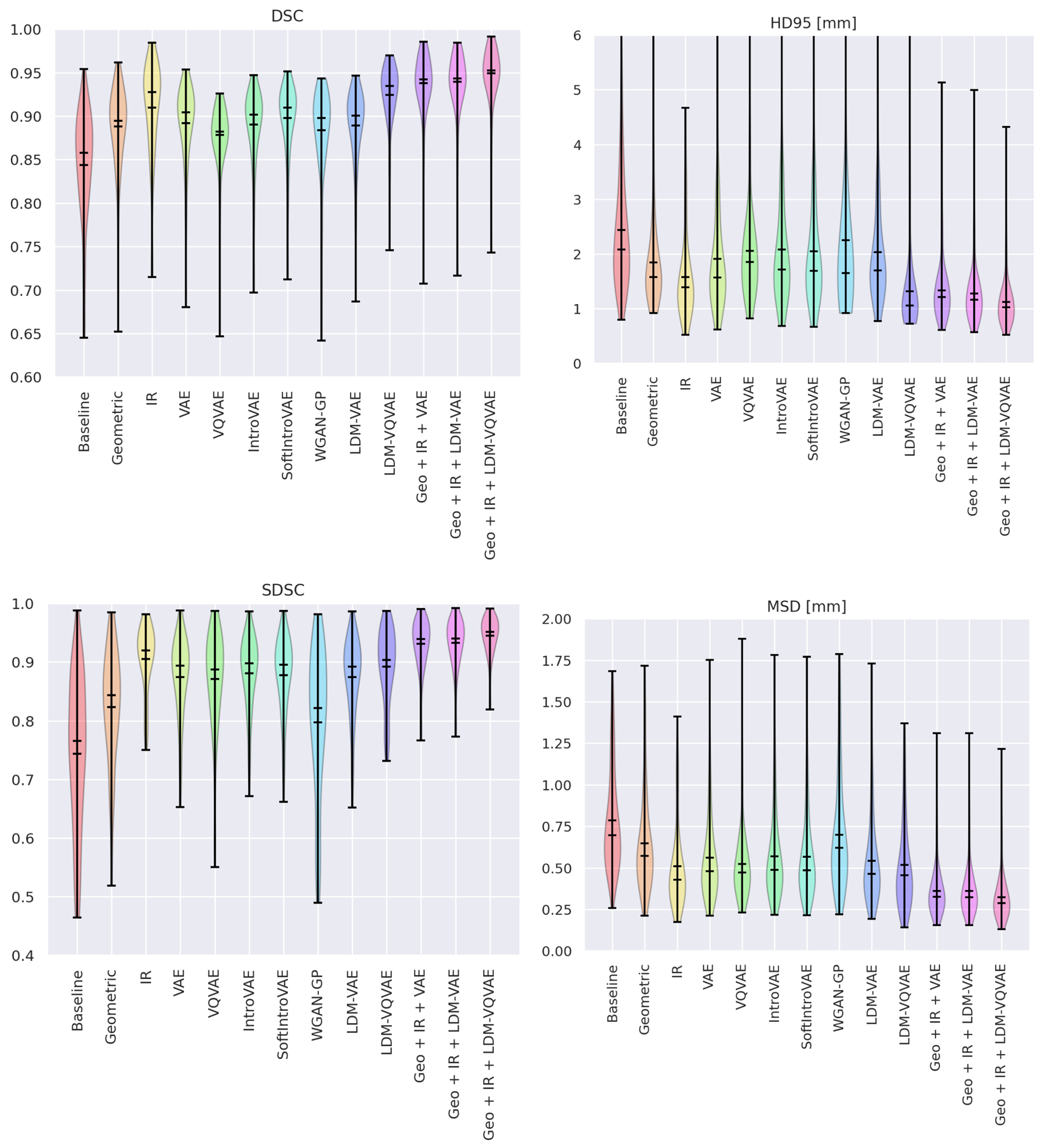}
    \caption{Quantitative comparison of different augmentation strategies and their combinations on the SkullBreak dataset. Note that all augmentation strategies improve over baseline, however, VAE- and WGAN-based methods perform significantly (p-value < 0.01) worse than image registration or latent diffusion models.}
    \label{fig:augmentation}
\end{figure*}

Another benchmark evaluates the different augmentation strategies. We compared the best-performing setup for all the augmentation methods. Thus, we have chosen all the generative methods that generated 64000 synthetic cases using the UDS sampling, the IR-based method with 64000 cases resulting from registration with the regularization coefficient equal to 12500, and the extreme geometric augmentation. It turned out that the best results were obtained by IR-based augmentation, VAE, LDM-VAE, and LDM-VQVAE. We decided to further extend the evaluation by combining the augmentation methods, therefore we randomly sampled and combined the datasets to create datasets containing 64000 samples generated by: (i) IR + VAE, (ii) IR + LDM-VAE, (iii) IR + LDM-VQVAE. Additionally, we enhanced these datasets by extreme geometric augmentation. This resulted in 13 experiments that are reported in Table~\ref{tab:augmentation} and presented in Figure~\ref{fig:augmentation}. From these experiments, it turned out that the most successful method was a combination of the geometric augmentation, IR, and LDM-VQVAE with a Dice coefficient above 0.94 for the SkullBreak dataset and 0.96 for the SkullFix dataset.

\begin{table*}[!htb]
\centering
\caption{Table quantitatively comparing the augmentation methods and their combinations using test subsets of both SkullBreak and SkullFix datasets.}
\renewcommand{\arraystretch}{1.0}
\footnotesize
\resizebox{0.99\textwidth}{!}{%
\begin{tabular}{lccccccccc}
\label{tab:augmentation}
Augmentation Method & \multicolumn{4}{c}{SkullBreak} & \multicolumn{4}{c}{SkullFix} \tabularnewline
\hline
\multicolumn{1}{c}{} & \multicolumn{1}{c}{DSC $\uparrow$}  &  \multicolumn{1}{c}{SDSC $\uparrow$} & \multicolumn{1}{c}{HD95 [mm] $\downarrow$} & \multicolumn{1}{c}{MSD [mm] $\downarrow$} & \multicolumn{1}{c}{DSC $\uparrow$} & \multicolumn{1}{c}{SDSC $\uparrow$} & \multicolumn{1}{c}{HD95 [mm] $\downarrow$} & \multicolumn{1}{c}{MSD [mm] $\downarrow$} \tabularnewline

Baseline (No Aug) & 0.843 & 0.743 & 2.438 & 0.787 & 0.892 & 0.881 & 1.882 & 0.554 
\tabularnewline

Geometric (Geo) & 0.888 & 0.824 & 1.842 & 0.648 & 0.904 & 0.922 & 1.572 & 0.482 
\tabularnewline

Image Registration (IR) & 0.910 & 0.905 & 1.574 & 0.511 & 0.934 & 0.952 & 1.321 & 0.362
\tabularnewline

VAE & 0.892 & 0.874 & 1.906 & 0.561 & 0.909 & 0.917 & 1.462 & 0.419 
\tabularnewline

VQVAE & 0.878 & 0.871 & 2.054 & 0.524 & 0.905 & 0.903 & 1.542 & 0.468 
\tabularnewline

IntroVAE & 0.890 & 0.881 & 2.084 & 0.570 & 0.908 & 0.915 & 1.483 & 0.421 
\tabularnewline

SoftIntroVAE & 0.898 & 0.877 & 2.048 & 0.566 & 0.911 & 0.916 & 1.479 & 0.423 
\tabularnewline

WGAN-GP & 0.883 & 0.797 & 2.248 & 0.700 & 0.897 & 0.894 & 1.672 & 0.512 
\tabularnewline

LDM-VAE & 0.889 & 0.874 & 2.032 & 0.541 & 0.912 & 0.911 & 1.578 & 0.417 
\tabularnewline

LDM-VQVAE & 0.924 & 0.891 & 1.321 & 0.518 & 0.931 & 0.942 & 1.381 & 0.349 
\tabularnewline

Geo + IR + VAE & 0.938 & 0.931 & 1.331 & 0.360 & 0.961 & 0.963 & 1.128 & 0.307 
\tabularnewline

Geo + IR + LDM-VAE & 0.940 & 0.933 & 1.276 & 0.359 & 0.964 & \textbf{0.968} & 1.084 & 0.305 
\tabularnewline

Geo + IR + LDM-VQVAE (Full Aug) & \textbf{0.949} & \textbf{0.945} & \textbf{1.127} & \textbf{0.324} & \textbf{0.968} & 0.964 & \textbf{1.072} & \textbf{0.298} 
\tabularnewline

\hline

\end{tabular}}

\end{table*}

\subsection{Comparison to the state-of-the-art}

To show the impact of the heavily augmented dataset on the reconstruction quality we compared it to other state-of-the-art methods~\cite{li2023towards}. The results are presented in Table~\ref{tab:sota}. The proposed method is currently the best performing one across all reported in the literature, improving over other state-of-the-art methods. It confirms the importance of data augmentation for tasks with a limited amount of available training data.

\begin{table*}[!htb]
\centering
\caption{Table presenting the best-performing augmentation settings when compared to other state-of-the-art methods for SkullFix and SkullBreak datasets. Please note that not all papers report results for both SkullBreak and SkullFix datasets and some reports results only to limited number of digits.}
\renewcommand{\arraystretch}{1.0}
\footnotesize
\resizebox{0.99\textwidth}{!}{%
\begin{tabular}{lccccccc}
\label{tab:sota}
Method & \multicolumn{3}{c}{SkullBreak} & \multicolumn{3}{c}{SkullFix} \tabularnewline
\hline
\multicolumn{1}{c}{} & \multicolumn{1}{c}{DSC $\uparrow$} & \multicolumn{1}{c}{BDSC $\uparrow$} & \multicolumn{1}{c}{HD95 [mm] $\downarrow$} & \multicolumn{1}{c}{DSC $\uparrow$} & \multicolumn{1}{c}{BDSC $\uparrow$} & \multicolumn{1}{c}{HD95 [mm] $\downarrow$} \tabularnewline

Proposed (Full Aug) & \textbf{0.95} & \textbf{0.95} & \textbf{1.13} & \textbf{0.97} & \textbf{0.96} & \textbf{1.07}
\tabularnewline

Mahdi et al.~\cite{mahdi2021u} & 0.8 & 0.81 & 3.42 & 0.9 & 0.93 & 3.59
\tabularnewline

Yang et al.~\cite{mahdi2021u} & 0.8 & 0.89 & 3.52 & N/A & N/A & N/A
\tabularnewline

Wodzinski et al.~\cite{wodzinski2021improving} & 0.9 & 0.93 & 1.60 & 0.93 & 0.95 & 1.48
\tabularnewline

Yu et al.~\cite{yu2021pca} & N/A & N/A & N/A & 0.8 & 0.77 & 3.68
\tabularnewline

Kroviakov et al.~\cite{kroviakov2021sparse} & N/A & N/A & N/A & 0.85 & 0.95 & 2.65
\tabularnewline

Pathak et al.~\cite{pathak2021cranial} & N/A & N/A & N/A & 0.9 & 0.95 & 2.02
\tabularnewline

Wodzinski et al.~\cite{wodzinski2023high} & 0.87 & 0.85 & 1.91 & 0.90 & 0.89 & 1.71
\tabularnewline

Friedrich et al.~\cite{friedrich2023point} & 0.87 & 0.89 & 2.45 & 0.90 & 0.93 & 1.69
\tabularnewline


Wodzinski et al.~\cite{wodzinski2022deep} & 0.89 & 0.93 & 1.60 & 0.93 & 0.95 & 1.47
\tabularnewline

Li et al.~\cite{li2023sparse} & N/A & N/A & N/A & 0.88 & 0.96 & 4.11
\tabularnewline

\hline

\end{tabular}}

\end{table*}

\subsection{Clinical Cases}

\begin{figure*}[!htb]
    \centering
    \includegraphics[width = 0.82\textwidth]{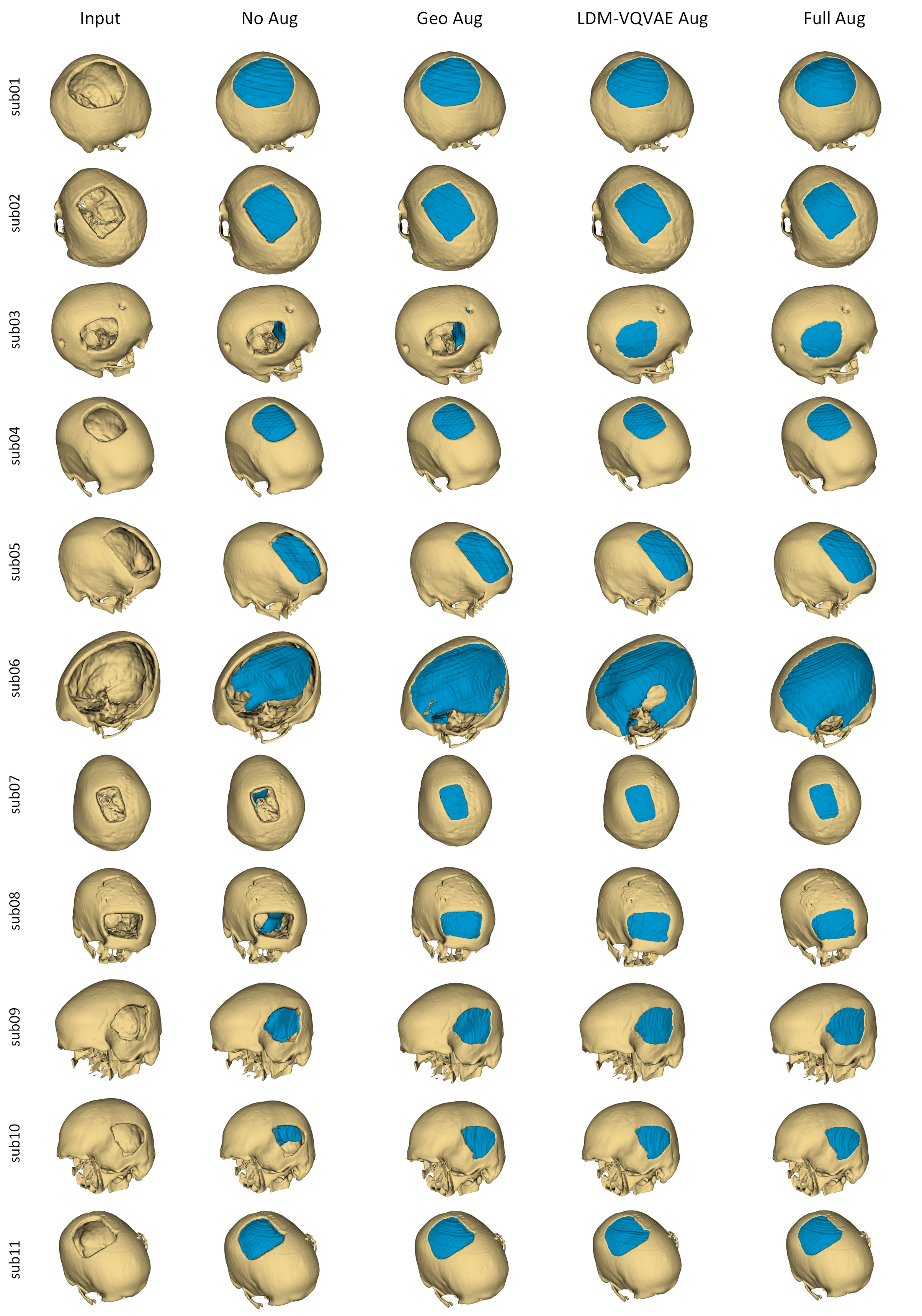}
    \caption{Influence of the augmentation on the real cranial defects from the AutoImplant challenge. Note that the combination of geometric, registration-based, and generative augmentation correctly reconstructs almost all the defects. Best viewed zoomed and in color.}
    \label{fig:aireal}
\end{figure*}

\begin{figure*}[!htb]
    \centering
    \includegraphics[width = 0.90\textwidth]{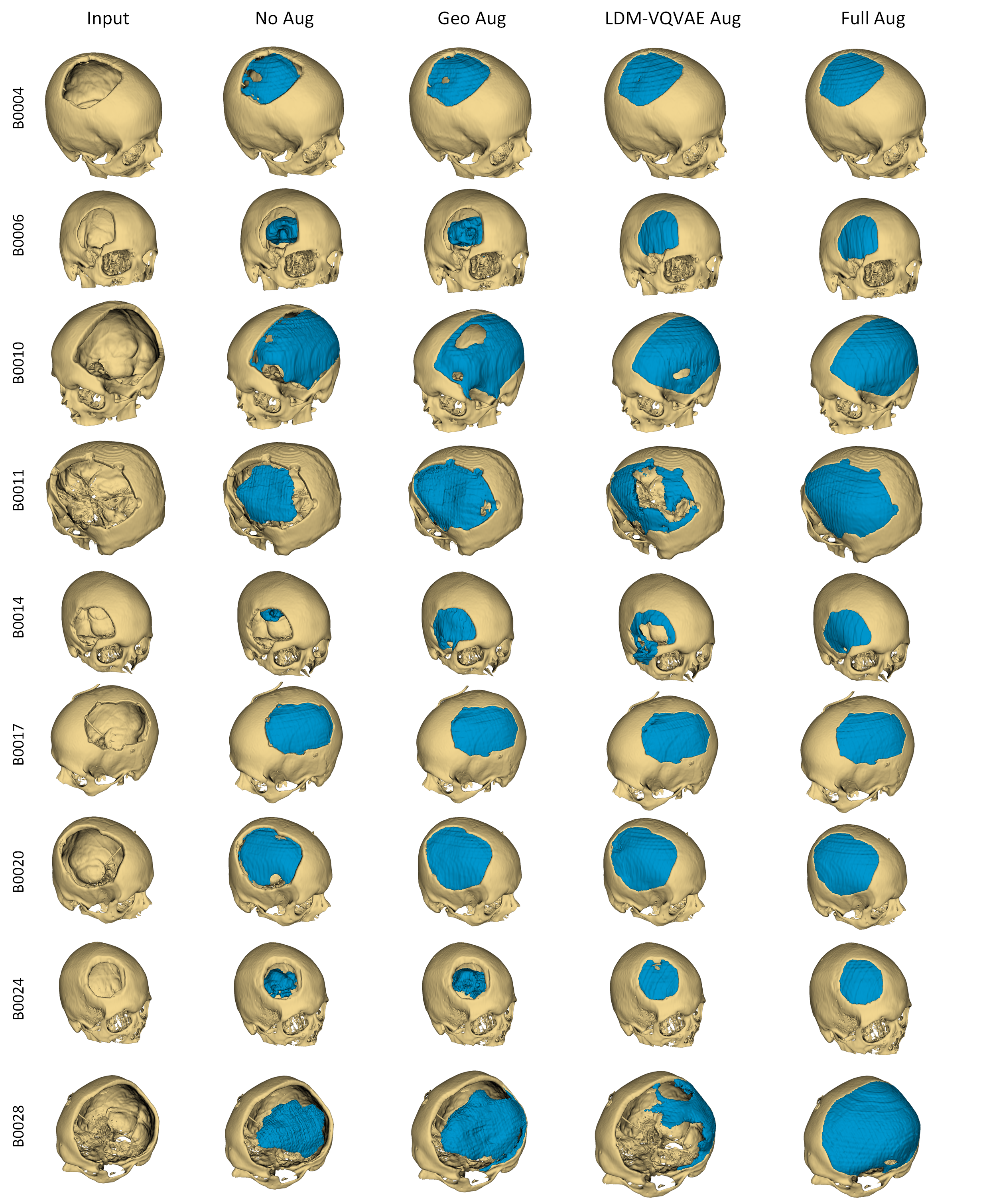}
    \caption{Influence of the augmentation on randomly selected real cranial defects from the MUG500 dataset. Note the difference when compared to Figure~\ref{fig:aireal}, the influence of the augmentation is significantly more influential for the MUG500 dataset (different distribution). Best viewed zoomed and in color.}
    \label{fig:mug500}
\end{figure*}

To further enhance the evaluation, we performed a qualitative evaluation using real clinical cases. Since for such cases, there is no ground truth available, we cannot calculate the quantitative values. Nevertheless, the results for models without the augmentation, with chosen augmentation strategies, and with the best possible augmentation strategy (GA + IR + LDM-VQVAE) are visually presented in Figure~\ref{fig:aireal} and Figure~\ref{fig:mug500} for the AutoImplant and MUG500 datasets respectively. It is visible that  augmentation is crucial for improving the generalizability to real clinical cases. The reconstruction network without any augmentation is unable to correctly reconstruct the majority of the real defects, while the heavily augmented method reasonably reconstructs all the cranial defects.

\section{Discussion}

The study confirms that data augmentation plays a crucial role in automatic cranial defect reconstruction. Since the ground truth is almost impossible to acquire for real clinical cases, the reconstruction algorithms need to be trained using synthetic defects. The proposed approach to extend the SkullBreak and SkullFix datasets by combining the geometric augmentation, deformable image registration, and latent diffusion model based on VQVAE improved the quantitative results for the SkullBreak dataset (when compared to the baseline) by more than 0.11, 1.30 mm, 0.19, 0.45 mm in terms of the DSC, HD95, SDSC, MSD respectively, resulting in average values equal to 0.949, 1.127 mm, 0.945, and 0.324 mm. The proposed approach is currently the most accurate solution among the state-of-the-art algorithms~(Table~\ref{tab:sota}). The strong augmentation plays a crucial role, especially for the quality of the surface reconstruction evaluated by the HD95 and MSD, the full augmentation strategy more than halved the baseline results (Table~\ref{tab:augmentation}). Interestingly, the combination of the LDM-VQVAE and IR-based augmentation is the best performing one. The reason behind this is connected with the fact that both these augmentation methods results in different distribution of synthetic cases, leading to significant improvement in generalizability. The IR-based registration results mostly in different skull shapes, with limited influence on the heterogeneity of the defect shapes. In contrast, the LDM-VQVAE augmentation results in heterogeneous defects, however, with limited influence on the skull heterogeneity. Combining both the methods and online geometric augmentation leads to a diverse training set, covering a wide range of possible scenarios.

Additionally, we confirmed the significant impact of the augmentation to improve the generalizability to real clinical cases. The method was able to correctly reconstruct almost all defects from the real clinical dataset while the same network, trained without augmentation, could not generalize the knowledge. Interestingly, the experiment confirms the necessity of performing evaluation using cases from different distributions. For the evaluation using test subsets of the SkullBreak and SkullFix datasets (Table~\ref{tab:augmentation}), the baseline achieves worse, however, still reasonable results. In contrast, for the real clinical cases, the method without heavy data augmentation does not operate correctly for the majority of the defective skulls (Figure~\ref{fig:mug500}).

The obtained results are accurate and robust (the worst DSC for the test set is above 0.70 for SkullBreak and 0.90 for SkullFix) basically solving the problem of automatic cranial reconstruction. It has significant clinical implications since defect reconstruction is a crucial step to model and manufacture personalized cranial implants.

Interesting observations can be made about the impact of synthetic dataset size. The deep models based on regularizing the latent space to enforce certain distribution quickly saturate and do not improve results beyond a certain threshold. It confirms that the generated samples are relatively homogeneous and similar to the ones the network was originally trained on. In contrast, the registration-based and generative models based on VQVAE where the latent space is represented by a discrete vector do not saturate that quickly, and increasing the dataset size further increases the network generalizability. Generating even a larger synthetic dataset and continuing to train the reconstruction model could result in even better results, however, it is beyond the available computational resources since it would require to also increase the reconstruction model expressability by increasing the model size or changing its architecture.

It is worth discussing the differences between VAE, VQVAE, LDM-VAE, and LDM-VQVAE in the context of data augmentation. It turned out that VAE and LDM-VAE achieved similar results, which is expected since the LDM based on the regularized VAE still uses the same decoder that was trained with KLD-based regularization. Denoising the randomly generated normal distributions results just in another normal distribution, basically without an impact on the heterogeneity of the generated data. The results are different for the VQVAE and LDM-VQVAE. Since the quality of samples generated by VQVAE depends strongly on the sampling function, generating the sampling prior by denoising normal distribution results in a better prior than using a deep sampler trained directly on the VQVAE training data. The differences between VQVAE and LDM-VQVAE are significant, since the VQVAE alone achieves worse performance than VAE/LDM-VAE, and the LDM-VQVAE outperforms both these methods (Table~\ref{tab:augmentation}).

The results obtained by VAE, IntroVAE, and SoftIntroVAE are similar, without considerable differences. The additional complexity of IntroVAE and SoftIntroVAE did not improve the results. The results obtained by WGAN-GP are among the worst ones achieved by the generative models, even though they still improve the outcomes compared to the baseline. What is more, training the WGAN-GP is the most computationally intensive, therefore we do not recommend using this model for generative data augmentation. Probably different variants of GANs could result in better outcomes, however, their computational cost for high-resolution volumetric data is beyond the available computational resources, even though they are significant.

Nevertheless, the approach based on heavy data augmentation has several limitations. The first and most important one is connected with the computational complexity. To benchmark all these models, perform the ablation studies, train the networks until convergence, and perform the evaluations, we had to use more than 100,000 A100 GPU hours (thanks to the PLGRID infrastructure). The commercial cost of such a benchmark could be significant, potentially discouraging people from using the generative approach, even though it has to be performed only once. The most computationally expensive part of the study was to train all the generative models, especially the WGAN-GP one. This was caused by problems with training stability, model collapse, and attempts to tune various hyperparameters, resulting in numerous experiments without any positive outcome. The positive thing is that the most stable in training were models based on VQVAE and LDMs which are also the best-performing ones. The computational resources could be potentially reduced by using super-resolution networks to upsample the generated representation at lower resolutions, however, this has to be done with caution because such an approach could result in destroying the fine details generated at higher resolutions. Another approach could be based on the networks dedicated to sparse data~\cite{li2023sparse}, however, this is topic for a further investigation. 

Another limitation is connected with the nature of IR. The IR-based augmentation can be applied only to datasets where the cases have similar morphology. Otherwise, the registration cannot be performed, limiting its applicability to only certain types of problems. Moreover, the IR-based approach requires tuning of the regularization coefficient for a particular problem that increases the computational complexity and enforces performing several experiments. Probably methods based on adaptive regularization coefficient could help to solve the problem~\cite{hoopes2021hypermorph,mok2021conditional}, however, it would require training a separate deep model for the IR-based augmentation.

Importantly, the study focuses on the impact of the augmentation strategies. We do not evaluate different reconstruction architectures for which the impact could be different. Nevertheless, repeating the experiments for various 3-D segmentation architectures would require enormous computational resources, beyond even the PLGRID supercomputing infrastructure capabilities.

In future work, we plan to explore the possibilities of performing similar augmentation using point cloud or surface mesh representation resulting in potentially lower computational complexity or combining both to achieve high-quality meshes directly from the volumetric representation~\cite{wickramasinghe2020voxel2mesh}. Moreover, since the desired output is a surface mesh, required to prepare the files for 3-D printing, the reconstruction in such a representation is closer to the final downstream task. Moreover, the generative models for point cloud or mesh representations could potentially generate the synthetic cases on the fly, resulting in online generative dataloaders, without the need to generate the samples before training the reconstruction model. What is more, we plan to directly evaluate the usability of LDM-VQVAE for cranial defect reconstruction since it enables inference-time augmentation, however, at the cost of increased reconstruction time. Another important research topic, currently under investigation, is the transfer of the reconstructed defect into the implantable model. The reconstructed defects cannot be used directly for 3-D printing because they have to be adapted to the desired material, the shape of the outer cavity, and the desired impact rim and thickness~\cite{fishman2024thickness}. Therefore, methods dedicated to transforming the reconstructed defect into personalized implants are highly desired.

It would be also interesting to perform a large-scale, multi-institution evaluation of the augmented and robust defect reconstruction model across several medical centers. We plan to start such an initiative in nearby future and any institution interested in automatic cranial defect reconstruction is warmly invited to participate.

To conclude, in this work, we evaluated several different types of data augmentation methods to improve the automatic cranial defect reconstruction. We have presented that a combination of geometric augmentation, image registration, and generative deep model based on VQVAE and latent diffusion, significantly improved the reconstruction quality and robustness. The method made it possible to perform reconstruction of real clinical cases, with a model trained purely on synthetic defects, confirming that the method is robust enough to generalize to previously unseen patients. The method obtained superior results that are close to clinical usability. This is a significant contribution to the field of artificial intelligence in computer-assisted neurosurgery.

\section*{Acknowledgements}

The project was funded by The National Centre for Research and Development, Poland under Lider Grant no: LIDER13/0038/2022 (DeepImplant). We gratefully acknowledge Polish HPC infrastructure PLGrid support within computational grants no. PLG/2023/016239 and PLG/2024/017079.

\section*{Disclosure of Interests} The authors have no competing interests to declare that are relevant to the content of this article.

\bibliographystyle{abbrv}
\bibliography{refs}
\end{document}